\documentclass[sigconf]{acmart}
\usepackage[utf8]{inputenc}
\usepackage{textcomp}
\usepackage{tikz}
\usepackage{pgfplots}
\usepackage{xcolor}
\usepackage{graphicx}

\usetikzlibrary{positioning,arrows.meta,decorations.pathmorphing}
\usepackage{amsthm}
\usepackage[nameinlink,capitalise]{cleveref}

\theoremstyle{plain}
\newtheorem{assump}{Assumption}
\usepackage{amsthm}

\theoremstyle{plain}

\newtheorem{lemma}{Lemma}
\newtheorem{proposition}{Proposition}

\theoremstyle{remark}
\newtheorem{remark}{Remark}
\usepackage{booktabs}

\usepackage{algorithm}      % 提供浮动体环境 \begin{algorithm}
\usepackage{algpseudocode}

\tikzset{>=Stealth}
\AtBeginDocument{%
  }

\setcopyright{none}
\begin{document}

\title{ CIP: A Plug-and-Play Causal Prompting Framework for Mitigating Hallucinations under Long-Context Noise}
\author{Qingsen Ma}
\authornote{These authors contributed equally to this work.}
\affiliation{%
  \institution{Beijing University of Posts and Telecommunications}
  \city{Beijing}
  \country{China}
}
\email{maqingsen@bupt.edu.cn}

\author{Dianyun Wang}
\authornotemark[1]
\affiliation{%
  \institution{Beijing University of Posts and Telecommunications}
  \city{Beijing}
  \country{China}
}

\author{Ran Jing}
\authornotemark[1]
\affiliation{%
  \institution{none}
  \city{Beijing}
  \country{China}
}

\author{Yujun Sun}
\affiliation{%
  \institution{Northwestern University}
  \city{Evanston, IL}
  \country{USA}
}

\author{Zhenbo Xu}
\authornote{Corresponding author.}
\affiliation{%
  \institution{Beijing University of Posts and Telecommunications}
  \city{Beijing}
  \country{China}
}

\renewcommand{\shortauthors}{Ma et al.}

\begin{abstract}
Large language models (LLMs) often hallucinate when processing long and noisy retrieval contexts, as they rely on spurious correlations rather than genuine causal relationships. To address this issue, we propose CIP, a lightweight, plug-and-play, causal inference framework that mitigates hallucinations at the input stage by constructing a causal relation sequence among entities, actions, and events and injecting it into the model prompt to steer reasoning toward causally relevant evidence. Through causal intervention and counterfactual reasoning, CIP suppresses non-causal reasoning paths, improving factual grounding and interpretability. Extensive evaluations across seven mainstream LLMs---including GPT-4o, Gemini 2.0-Flash, and Llama-3.1---show that CIP consistently enhances reasoning quality and reliability, achieving \textbf{+2.6} points in Attributable Rate (AR), +0.38 in Causal Consistency Score (CCS), and a fourfold increase in effective information density. Moreover, API-level profiling demonstrates that CIP’s causal pre-analysis accelerates contextual understanding and reduces end-to-end response latency by up to \textbf{55.1\%}. These results suggest that causal reasoning may serve as a promising paradigm for enhancing the explainability, stability, and efficiency of large language models.
\end{abstract}


% \ccsdesc[500]{Computing methodologies~Natural language processing}
% \ccsdesc[500]{Computing methodologies~Causal reasoning and diagnostics}
% \ccsdesc[300]{Computing methodologies~Knowledge representation and reasoning}
% \ccsdesc[300]{Information systems~Information extraction}
% \ccsdesc[300]{Information systems~Question answering}



\maketitle

\section{Introduction}
The rise of large language models (LLMs) is reshaping the paradigm of information processing and human-computer interaction \cite{Kojima2023ZeroShot, Wei2022ChainOfThought}, demonstrating unprecedented potential in knowledge-intensive fields such as medical diagnosis \cite{Alber2025Medical} and legal document analysis \cite{Ma2024CausalSurvey}. However, despite these models' growing capabilities, a fundamental challenge—hallucinations—remains a major obstacle to their reliable application \cite{Huang2025Survey, Banerjee2024hallucinate}. When presented with lengthy and complex documents, models tend to generate content that is inconsistent with the facts or even fabricated. In critical scenarios like medicine and law, such confident, erroneous outputs can not only lead to catastrophic consequences but also severely erode user trust in AI systems, becoming a key bottleneck hindering their widespread deployment.

\begin{figure}
    \centering
    \includegraphics[width=1\linewidth]{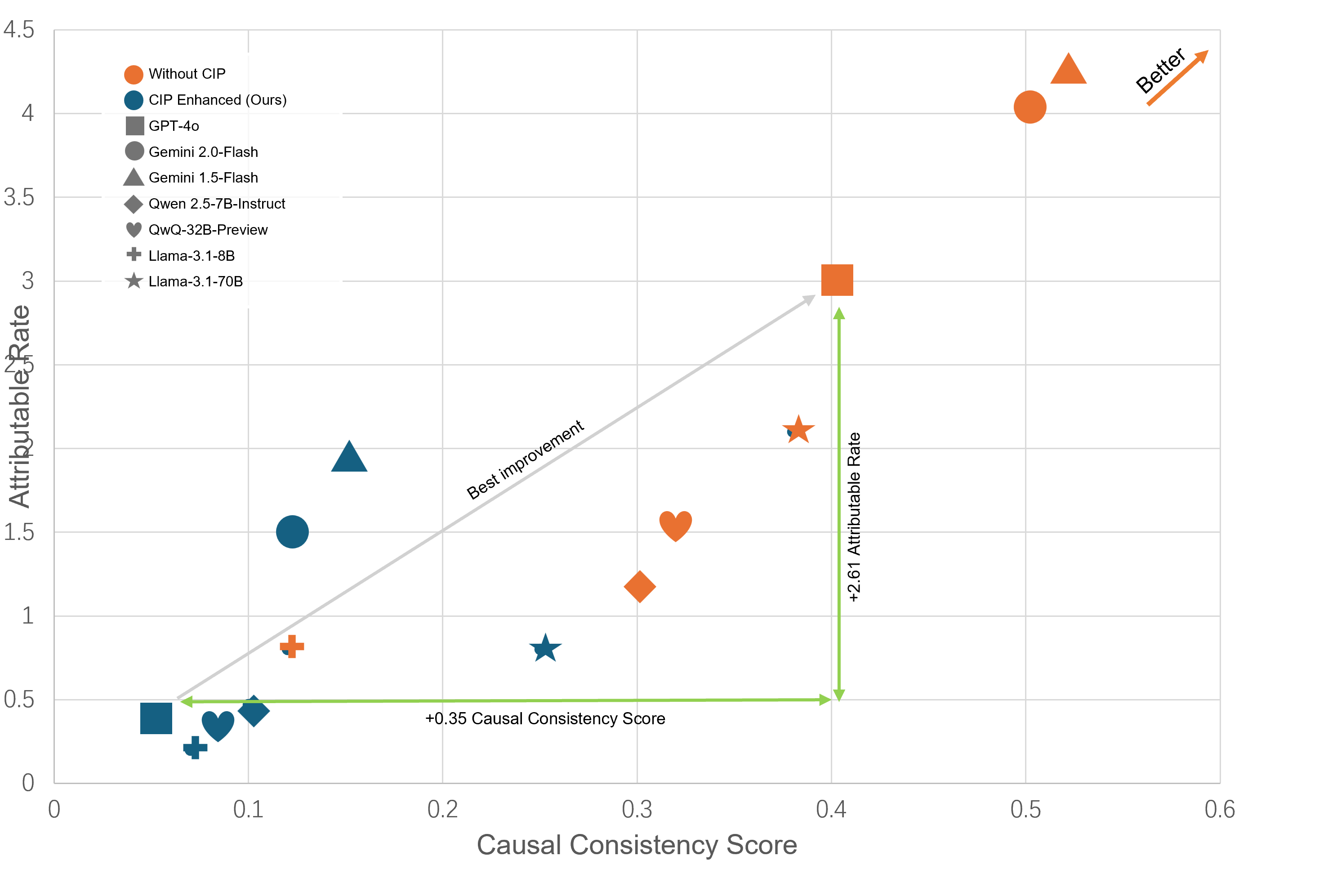}
    \caption{Scatter plot showing improved Attributable Rate and Causal Consistency Score with CIP enhancement.}
    \label{fig:placeholder}
\end{figure}
To mitigate hallucinations, Retrieval-Augmented Generation (RAG) has been widely adopted to inject external factual knowledge into LLMs by retrieving relevant documents. However, RAG acts primarily as an “information transporter” rather than an “information discriminator,” often delivering noisy, irrelevant, or contradictory content—especially when processing long documents. This indiscriminate injection of retrieval noise exacerbates the very issue it aims to solve, as LLMs are left to reason over chaotic contexts without guidance on causal relevance, thereby reinforcing misinterpretation and hallucination.

Hallucinations in large language models (LLMs) are not merely the byproduct of insufficient retrieval or missing factual knowledge, but manifestations of deeper systemic flaws in how models generate and evaluate information. As argued by \cite{Kalai2025Why}, hallucinations arise from the joint effect of statistical pressure and evaluation bias. During pre-training, LLMs are optimized to maximize linguistic fluency across massive corpora, learning to reproduce “credible fallacies” without distinguishing truth from plausibility. This tendency becomes especially severe when dealing with long documents containing sparse factual anchors, where the model’s statistical prior dominates reasoning. Simultaneously, as \cite{Damani2025Beyond} highlights, the dominant post-training and evaluation paradigms penalize expressions of uncertainty. Models are thus incentivized to make confident guesses rather than to abstain or indicate ambiguity when faced with noisy, contradictory, or incomplete evidence. These forces jointly drive the model to fabricate details to maintain narrative coherence.

Current methods mainly attempt to mitigate hallucinations by supplying additional external information, thereby improving retrieval fidelity. However, this approach burdens the model with noisy or spurious correlations within the retrieved content, which can easily mislead reasoning and amplify hallucinations.In contrast, incorporating causal logic aims to enhance reasoning fidelity — enabling large language models to assess the causal relevance of retrieved evidence rather than treating all information as equally valid. By grounding inference in causal consistency instead of surface-level statistical alignment, models can move from information accumulation toward reasoned understanding, fundamentally reducing hallucinations rather than merely masking them.

Causal Inference Plugin (CIP) is a lightweight, plug-and-play prompting framework that restructures raw retrieval outputs into explicit causal relation chains among entities, events, and actions. Drawing on causal inference principles, CIP focuses on extracting, filtering, and representing only causally relevant evidence. Through this process, it reformulates retrieved content into structured causal graphs or chains that serve as inputs for reasoning, enabling large language models to process information according to causal dependencies rather than surface-level correlations Our CIP has three key advantages.
\textbf{(1) Improving Output Quality:} By extracting causal knowledge through CIP, we not only substantially reduce the hallucination rate but also achieve a marked enhancement in overall output quality. Specifically, CIP enables (a) \textit{enhanced logical rigor}, ensuring the model’s responses follow clear causal reasoning; (b) \textit{improved information density}, effectively eliminating redundant or fragmented expressions and thereby increasing the “effective information density” of the generated text; and (c) \textit{enhanced long-text comprehension capabilities}, as demonstrated by superior performance on multi-document and long-context understanding tasks. 
\textbf{(2) Plug-and-Play:} CIP is designed as a modular, input-level auxiliary unit that seamlessly integrates with existing mainstream LLMs without requiring parameter modifications. We validated its compatibility and effectiveness across seven industry-leading models, including \cite{Grattafiori2024Llama3, Hurst2024GPT4o, Jiang2023Mistral, Baichuan2023, Taori2023Alpaca, Chiang2023Vicuna, Anand2023GPT4All}, achieving consistent performance gains. 
\textbf{(3) Resource-Efficient:} CIP is highly efficient in both training and deployment. Built upon a 7B-parameter language model and fine-tuned using LoRa technology \cite{Hu2022LoRA}, it avoids the need for complex reinforcement learning procedures or auxiliary reward/reference models, thereby substantially reducing computational and financial costs while maintaining strong adaptability and scalability.

The main contribution of this paper is the proposal and implementation of a novel and efficient Causal Inference Plugin (CIP). By addressing the source of information processing, it provides a new perspective for addressing the hallucination problem of large language models in long document understanding. In the following sections, we will elaborate on the theoretical foundations and model architecture of CIP, demonstrate its empirical results that outperform existing methods on multiple benchmarks \cite{Li2023HaluEval, Wang2024CausalBench, Yu2025CausalEval, Jin2023CLadder}, and further explore the broad prospects of causal inference in building more reliable and trustworthy AI systems."

\section{Related Work}

\subsection{Hallucinations in Large Language Models}
Recent work has explored the phenomenon of hallucinations in large language models (LLMs), attributing them to statistical pressures within the training process and evaluation mechanisms that prioritize fluency over factual accuracy \cite{Kalai2025Why, Huang2025Survey}. Kalai et al. \cite{Kalai2025Why} argue that these hallucinations arise because LLMs are optimized to perform well on benchmark tests that penalize uncertainty, encouraging models to generate plausible-sounding responses rather than admit uncertainty. The authors suggest that this "epidemic" of penalizing uncertain responses leads to the fabrication of details to fill in logical gaps in model outputs, even when evidence is noisy or conflicting \cite{Damani2025Beyond}. Their work emphasizes the need for a shift in the evaluation framework to improve the reliability and trustworthiness of AI systems, proposing the modification of existing benchmarks rather than the introduction of new hallucination-specific evaluations \cite{Gao2024bHarness}.

\subsection{Long-Context Comprehension Challenges}

In addition to hallucinations, challenges related to long-context understanding in LLMs have been highlighted in recent studies \cite{Liu2023Lost}. Liu et al. \cite{Liu2023Lost} investigate the performance of language models on tasks requiring the identification of relevant information within long input contexts. They observe that model performance significantly degrades when the position of relevant information within long contexts is changed, even for models explicitly designed for long-context processing. Their findings suggest that while current LLMs can process long inputs, they struggle to effectively utilize information located in the middle of the context. This degradation in performance further underscores the need for improved methods to handle long-term dependencies in document-level understanding, particularly in multi-document tasks like question answering and key-value retrieval.

\begin{figure*}
    \centering
    \includegraphics[width=1\linewidth]{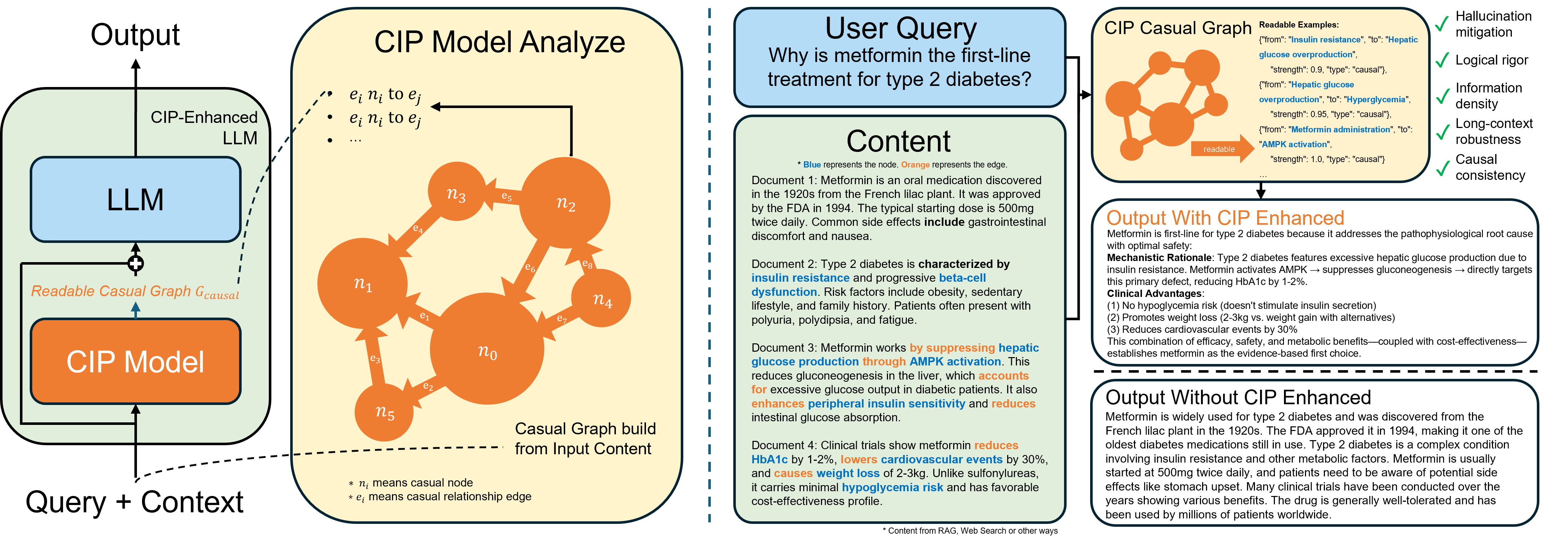}
    \caption{Overview of the CIP (Causal Inference-based Prompting) framework. The \textbf{content} refers to long-form or multi-document inputs that contain complex factual dependencies and potential noise, such as medical summaries, research reports, or policy documents.  CIP directly analyzes the given content . It performs a causal pre-analysis to identify entities, events, and their cause–effect relations, constructing a structured causal graph that is integrated into the prompt. This process does not alter the original content but enhances the LLM’s reasoning by guiding attention toward causally relevant information and suppressing non-causal context. The highlighted blue and yellow regions illustrate how CIP preserves essential causal cues while filtering irrelevant information, resulting in outputs with stronger logical coherence, higher factual faithfulness, and reduced hallucinations. The example demonstrates CIP’s effectiveness in a multi-document medical summarization scenario.}
    \label{fig:placeholder}
\end{figure*}

\section{Background: Causal Inference as a Remedy for Hallucination}

Large language models (LLMs) often produce hallucinations—plausible yet incorrect statements—because they rely on statistical correlations rather than causal understanding \cite{Kalai2025Why, Huang2025Survey}. During pre-training, these models learn linguistic regularities from massive text corpora but do not acquire mechanisms to distinguish between \textit{what is likely to appear together} and \textit{what actually causes what}. As a result, when faced with incomplete or noisy information, they tend to generate text that is coherent but not necessarily factual, driven by the statistical pressure to maintain fluency. This phenomenon is particularly severe in long-document or multi-source reasoning tasks, where superficial co-occurrence patterns dominate over true causal logic \cite{Ji2023Survey, Liu2023Lost}.

Causal inference (CI) offers a principled framework to address this fundamental limitation. Rather than modeling correlations, CI focuses on uncovering the \textit{underlying mechanisms} that produce observable data \cite{Hernan2020Causal}. In the causal view, the world is not a collection of co-occurring facts but a system of interacting variables connected by directed causal relations. This distinction allows models to reason about how changes in one variable propagate to others—something purely statistical systems cannot capture.

To illustrate, consider a medical diagnosis scenario. A patient exhibits \textit{fever}, \textit{cough}, and \textit{chest pain}. A correlation-based model might predict “COVID-19” simply because these symptoms frequently co-occur with that label in the training data. However, causal reasoning identifies that “viral infection” causes “inflammation,” which leads to both “fever” and “cough,” while “chest pain” may result from an independent condition. By explicitly representing these causal chains, the model can discern genuine causal relevance from mere coincidence, avoiding spurious conclusions. This principle directly parallels how CI can suppress hallucinations in language models: by forcing reasoning to follow valid causal pathways instead of arbitrary token associations.

In text understanding, applying CI involves transforming unstructured language into structured causal graphs or event chains \cite{Luo2025CausalGraphs, Fu2025Unveiling}. These representations encode entities, actions, and relationships in a way that mirrors the underlying logical structure of the text. When integrated with LLMs, they act as a “reasoning scaffold,” constraining generation to remain faithful to causal dependencies derived from the source material \cite{Kiciman2023Causal}. This reduces the model’s incentive to speculate, particularly when retrieved evidence is sparse or contradictory, thus addressing hallucinations at their root.

In essence, causal inference reframes language generation as a process of reasoning over \textit{cause and effect}, rather than surface-level co-occurrence. By embedding causal structure into the input or reasoning stage, we transform LLMs from statistical predictors into interpretable, causally grounded reasoners. This paradigm shift—from correlation to causation—forms the theoretical foundation of our proposed Causal Inference Plugin (CIP).

\section{Method}

\subsection{Theory: CIP for Suppressing Hallucinations}

\textbf{Notation and Setup.}  
Consider a structural causal model (SCM) \( \mathcal{M} \) with variables: query \( Q \), fact variables \( F \) (verifiable evidence from documents), spurious/irrelevant variables \( S \) (redundant details, outdated or contradictory content), observed context \( X := \phi(F,S,U_X) \), true answer \( Y^\star := g(F,Q,U_Y) \), with \( U_X, U_Y \) independent exogenous noise. Let the LLM output be \( \widehat{Y} \). We formalize hallucination as the event that the predicted answer \( \widehat{Y} \) is not in the admissible answer set \( \mathcal{A}(F,Q) \), where \( \mathcal{A}(F,Q) \) is the admissible answer set induced by facts and query.We further introduce a causal intervention representation \( R := \tau(X, Q) \), 
which denotes the causally refined input obtained by applying the CIP transformation \( \tau(\cdot) \) 
to the observed context \( X \) and query \( Q \); \( R \) serves as the causally sufficient and deconfounded version of \( X \).

The hallucination risk under information \( W \) (where \( W = X \) or \( W = R \)) is defined as follows:

\[
\mathcal{R}^\star(W) = \inf_{\pi} \mathbb{E}\!\left[ 1_{\{\pi(W) \notin \mathcal{A}(F,Q)\}} \right] = \mathbb{E}\!\left[ 1 - \max_{y \in \mathcal{A}(F,Q)} P(Y^\star = y \mid W) \right].
\]

This formula encapsulates the risk of hallucination for the model given certain information \( W \).

First, \textbf{factual sufficiency} ensures that the information contained in the fact variables \( F \) is a subset of the information encoded in \( R \), and that the true answer \( Y^\star \) is independent of the observed context \( X \) given \( R \) and the query \( Q \), making \( R \) sufficient for \( Y^\star \). Second, \textbf{deconfounding} guarantees that \( R \) is independent of the spurious or irrelevant variables \( S \), conditioned on the fact variables \( F \) and the query \( Q \), thus ensuring that \( R \) does not carry any spurious dependence. Finally, \textbf{identifiability} states that the true answer \( Y^\star \) can be identified from the intervention \( R \), meaning that the distribution of \( Y^\star \) given a do-operation on \( Q \) and \( F \) is identifiable from \( R \). Therefore, \( R \) encodes the causal structure while filtering out irrelevant variables \( S \). 

\textbf{Preliminaries.}  
Let $\mathcal{P}_{\mathrm{shift}}$ denote the family of distributions obtained from the original data distribution by arbitrary but admissible shifts in spurious or irrelevant variables $S$.  
Let $\mathcal{R}^{\mathrm{rob}}(W)$ denote the \emph{robust hallucination risk} of a representation $W$, defined as 
\[
\mathcal{R}^{\mathrm{rob}}(W) = \sup_{P' \in \mathcal{P}_{\mathrm{shift}}} \mathcal{R}^\star_{P'}(W),
\]
i.e., the worst-case hallucination risk under distributional perturbations.  
Let $L_\tau$ be the Lipschitz constant of the causal transformation $\tau(\cdot)$ used to map $X$ to $R = \tau(X, Q)$.

\textbf{Assumptions (C1–C3).}  
We assume the following conditions hold:
\begin{itemize}
    \item \textbf{(C1) Factual sufficiency:} $Y^\star \!\perp\!\! X \mid (R,Q)$ and the information in $(F,Q)$ is contained in $R$. 
    \item \textbf{(C2) Deconfounding:} $R \!\perp\!\! S \mid (F,Q)$, i.e., $R$ is independent of spurious variables once conditioned on true facts and query.
    \item \textbf{(C3) Identifiability:} $P(Y^\star \mid \mathrm{do}(F,Q)) = P(Y^\star \mid R,Q)$, ensuring causal sufficiency of $R$ for predicting $Y^\star$.
\end{itemize}

\textbf{Lemma 4.1 (Causal Invariance).}  
Under conditions (C1)–(C3), for any $P' \in \mathcal{P}_{\mathrm{shift}}$, the conditional distribution of the true answer given $R$ and $Q$ is invariant to distribution shifts:
\[
P'(Y^\star \mid R,Q) = P(Y^\star \mid R,Q).
\]
Thus, $R$ preserves causal invariants across domains.

\textbf{Theorem 4.2 (Upstream Intervention Principle).}  
Under (C1)–(C3), the robust hallucination risk of the causally refined input $R$ is no greater than that of the original observed context $X$:
\[
\mathcal{R}^{\mathrm{rob}}(R) \le \mathcal{R}^{\mathrm{rob}}(X),
\]
with strict inequality if there exist spurious causal paths $S \!\rightarrow\! X \!\rightarrow\! Y^\star$.

\textbf{Corollary A (Logical Rigor).}  
Because $R$ encodes consistent causal relations among facts,
\[
\mathbb{P}(\text{logical violation}) = \mathcal{R}^\star(R) \le \mathcal{R}^{\mathrm{rob}}(X),
\]
ensuring stronger logical coherence in generated outputs.

\textbf{Corollary B (Information Density).}  
Define the \emph{Effective Information Density} (EID) of a representation $W$ as
\[
\mathrm{EID}(W) = \frac{I(F;W \mid Q)}{\mathbb{E}[|W|]}.
\]
Since CIP removes $S$ while preserving $F$, we have $I(F;R \mid Q) \ge I(F;X \mid Q)$ while $\mathbb{E}[|R|] < \mathbb{E}[|X|]$, implying $\mathrm{EID}(R) > \mathrm{EID}(X)$.  
Thus, CIP yields a more fact-dense input, reducing redundancy and sparsity.

\textbf{Corollary C (Long-Document Generalization).}  
For a decoder class $\mathcal{H}$ with Lipschitz constant $L$, the Rademacher complexity satisfies
\[
\mathfrak{R}_n(\ell \circ \mathcal{H} \circ \tau) 
\le L_\tau \, \mathfrak{R}_n(\ell \circ \mathcal{H}),
\]
where $L_\tau \ll 1$ because $\tau$ compresses $X$ into the low-variance causal representation $R$.  
Hence, CIP reduces the generalization gap and improves long-context reasoning robustness.

\textbf{Distance Bound.}  
By Pinsker’s inequality,
\[
\mathcal{R}^\star(W)
\le \tfrac{1}{2} \mathbb{E}_{(F,Q)} 
\!\left[\! \sqrt{\mathrm{KL}\!\left(P_W \Vert P_{F,Q}\right)} \, \right].
\]
For $R$ where $P_R = P_{F,Q}$, $\mathcal{R}^\star(R) = 0$ in the ideal case, or $\le \sqrt{\epsilon}/2$ for finite error $\epsilon$.  
In contrast, for $X$, divergence from irrelevant variables $S$ increases the hallucination risk.

In summary:  
CIP performs causal upstream intervention, converting noisy context \( X \) into a causal-sufficient representation \( R \). This process ensures robust suppression of hallucination risk under distribution shifts, strengthens logical consistency, increases effective information density, and improves generalization on long-document tasks. This approach formally supports the "river upstream" theory, which argues that intervening directly in the input causal structure reduces hallucinations at their source, rather than trying to patch them downstream.

\subsection{Causal Web Dependency Pre-Identification}
\label{subsec:causal-web-preid}
Conventional WebTool–enhanced pipelines trigger retrieval \emph{reactively} during decoding: when the model encounters a knowledge gap, it halts generation, issues a query, and waits for results before resuming \cite{Yao2023ReAct,Schick2024Toolformer}. This design causes idle token latency, underutilized accelerators, and compounded delays in long- or multi-document settings where multiple sequential queries are triggered.

\noindent\textbf{CIP as a proactive scheduler.}
Before decoding, the Causal Inference Plugin (CIP) analyzes the input’s causal structure and decomposes the reasoning chain into \emph{endogenous} nodes (resolvable from internal facts $F$) and \emph{exogenous} nodes (requiring external evidence $S_{\text{ext}}$). Retrieval-critical nodes are then dispatched to the WebTool \emph{in parallel} \emph{prior} to token generation, so decoding proceeds uninterrupted once it starts.
\paragraph{Example (post-training trial query)}
\emph{Question:} ``The 2025 August trial XYZ-CV studied the effect of metformin vs GLP-1 agonist Z in reducing major adverse cardiovascular events (MACE). Based on all documents, how do their cardiovascular benefits compare?''

CIP constructs a causal graph and determines that only two exogenous relations must be retrieved externally:

 $\text{GLP-1 agonist Z (trial XYZ-CV)} \rightarrow \text{MACE outcomes}$,
 
 $\text{Metformin (trial XYZ-CV)} \rightarrow \text{MACE outcomes}$.

CIP issues both web queries simultaneously:
- “2025 XYZ-CV trial GLP-1 agonist Z cardiovascular outcomes”  
- “2025 XYZ-CV trial metformin cardiovascular outcomes”  

After fetching and integrating the trial results (e.g.\ hazard ratios, confidence intervals, patient subgroups), the model can reason causally:
> “In the XYZ-CV trial, GLP-1 agonist Z reduced MACE by 15% (HR 0.85 [0.78–0.93]), while metformin achieved a 10% reduction (HR 0.90 [0.82–0.99]). The greater benefit of Z is likely mediated via weight loss, improved lipid profile, and anti-inflammatory effects. Metformin contributes via AMPK activation and modest lipid lowering. Importantly, hypoglycemia risk remained lower in the metformin arm.”

Thus the LLM generates a coherent comparative answer in one pass—no mid-generation retrieval interruptions.
\paragraph{Benefits}
This upstream, causally guided scheduling yields:
\textbf{Lower latency}: eliminates decode-time waiting by resolving $S_{\text{ext}}$ before generation;\textbf{Higher efficiency}: overlaps retrieval with input parsing, improving hardware utilization;\textbf{Better evidence quality}: targets are derived from the causal graph rather than shallow correlations, increasing effective information density and logical alignment.

In sum, CIP converts retrieval from a reactive, token-level decision into a proactive, input-level plan, improving both reliability (via causal grounding) and system efficiency (via latency reduction).

\subsection{LoRA-based Causal Fine-Tuning}

We fine-tune only the CIP plugin (7B causal extractor) with LoRA, keeping the upstream LLM frozen. CIP’s output is used as prompt augmentation, and no base-LLM weights are updated.

LoRA introduces lightweight low-rank adapters inside CIP, effectively correcting systematic causal interpretation biases (e.g., direction inversion, missing causal links) while remaining computationally efficient. This yields consistent improvements in causal directionality and completeness.

\paragraph{Structured Supervision.}
A structured dataset is critical for causal alignment. Our semi-automatic pipeline ensures data quality through:
GPT-4o Generation – broad coverage of causal forms.Knowledge Distillation – transferring extraction patterns to CIP’s smaller backbone.Metric Filtering – retaining only pairs with causal/semantic scores >0.9.Human Verification – final correction to ensure causal intent.
\paragraph{Results.}
LoRA fine-tuning of CIP improves causal consistency +(6.3\%), semantic coherence (+5.8\%), and structural validity (+4.9\%).
\paragraph{Summary.}
LoRA-based fine-tuning of CIP efficiently corrects causal-reasoning bias while keeping the LLM frozen. Dataset structuring is essential to prevent spurious causality and maintain reasoning fidelity.

\begin{table*}[h!]
\centering
\resizebox{\textwidth}{!}{
\begin{tabular}{|l|c|c||c|c||c|c||c|c|c|}
\hline
\textbf{Model Variant} & \textbf{Causal AC} & \textbf{Direct AC} & \textbf{Causal CCS} & \textbf{Direct CCS} & \textbf{Cohen’s d (AR)} & \textbf{Cohen’s d (CCS)} & \textbf{p-value} & \textbf{$\Delta$ AC} & \textbf{$\Delta$ CCS}  \\
\hline
GPT-4o  & 2.96 & 0.35 & 0.40 & 0.05   & 0.48 & 0.42 & < $10^{-3}$ & \textbf{+2.61} & \textbf{+0.35} \\
Gemini-2.0-Flash  & 4.02 & 1.48 & 0.50 & 0.12  & 0.25 & 0.28 & < $10^{-3}$ & \textbf{+2.54} & \textbf{+0.38} \\
Gemini-1.5-Flash  & 4.25 & 1.92 & 0.52 & 0.15  & 0.21 & 0.23 & < $10^{-3}$ & \textbf{+2.33} & \textbf{+0.37} \\
Qwen-2.5-7B-Instruct  & 1.16 & 0.47 & 0.30 & 0.10 & 0.16 & 0.17 & 0.001 & \textbf{+0.69} & \textbf{+0.20} \\
Qwen-32B-Preview  & 1.50 & 0.35 & 0.32 & 0.08   & 0.23 & 0.22 & < $10^{-3}$ & \textbf{+1.15} & \textbf{+0.24} \\
Llama-3.1-8B  & 0.80 & 0.20 & 0.25 & 0.07   & 0.18 & 0.16 & 0.002 & \textbf{+0.60} & \textbf{+0.18} \\
Llama-3.1-70B  & 2.10 & 0.80 & 0.38 & 0.12  & 0.35 & 0.33 & < $10^{-3}$ & \textbf{+1.30} & \textbf{+0.26} \\
\hline
\end{tabular}
}
\caption{Unified Comparison of Attributable Count (AC), Attributable Rate (AR), and Causal Consistency Score (CCS) under Causal and Direct Prompting.}
\end{table*}

\subsection{Causal Reasoning Dataset Construction}

The dataset construction process is essential for training the causal inference model. To ensure high-quality, meaningful causal relationships, we follow a structured extraction process from both knowledge and dialogue sources. The goal is to identify and capture various forms of causal, attribute, and factual relationships, which serve as the foundational elements for our model’s reasoning capabilities.

\subsubsection{Causal Relationship Extraction}

The core of our dataset involves the extraction of three distinct types of relationships from the provided knowledge and dialogue:

1. \textbf{Direct Causal Relationships}: These involve direct cause-and-effect relationships, expressed as "X causes Y" or "X leads to Y." For example, "Heavy rainfall causes flooding" or "Increased demand leads to higher prices."
2. \textbf{Attribute Relationships}: These describe the inherent properties of entities, such as "X has property Y" or "X is Y." For instance, "The car has a blue color" or "Alice is a doctor."
3. \textbf{Factual Relationships}: These encompass factual connections, such as "X starred in Y" or "X wrote Y." An example would be "Leonardo DiCaprio starred in Titanic" or "J.K. Rowling wrote Harry Potter."

The extraction process follows a methodical approach to identify these relationships from both structured knowledge sources and conversational dialogue. The relationships are encoded in a JSON format for ease of processing and integration into our causal inference framework.

\subsubsection{Output Format}

The relationships are structured in the following JSON format to facilitate further processing:

\begin{verbatim}
{
  "nodes": ["entity1", "entity2"],
  "edges": [
    {
      "from": "cause", "to": "effect", 
      "strength": 0.9, "type": "causal"
    }
  ]
}
\end{verbatim}

Here, nodes represent entities or concepts, while edges define the relationships between them. Each edge is annotated with a strength value that quantifies the degree of the relationship (with a value between 0 and 1), and a type field that specifies whether the relationship is causal, attribute-based, or factual.

\subsubsection{Human Annotation}

Once the raw data is extracted, the dataset undergoes a thorough manual review process. We examine a sample of 3000 instances to ensure that the relationships are accurately captured and appropriately classified. This manual review step is crucial for ensuring the quality and relevance of the data used to fine-tune the causal inference model.

This approach provides a scalable and efficient method for integrating causal reasoning into large language models, ensuring that the model can handle long-context scenarios and complex multi-document tasks.

\section{Experiment}

\subsection{Causal Inference Suppresses Hallucinations}

In this section, we present the rigorous evaluation of causal prompting applied to several state-of-the-art language models, including GPT-4o, Gemini 2.0-Flash, Gemini 1.5-Flash, Qwen/Qwen 2.5-7B-Instruct, Qwen/QwQ-32B-Preview, and Llama-3.1 variants. We focus on demonstrating that causal prompting significantly reduces hallucinations, improves the logical consistency of model outputs, and enhances the effective information density in generated responses. The experimental setup and results are discussed below.
\subsubsection{Experimental Setup}
We applied causal prompting to different model variants, including both smaller models (e.g., \textbf{Qwen-7B}) and larger models (e.g., \textbf{Llama-70B}). The models were evaluated using a set of 800 carefully curated samples, ensuring rigorous testing conditions.

Unlike conventional evaluations of large language models that focus primarily on surface-level metrics—such as BLEU, ROUGE, or perplexity—our study seeks to measure a model’s \textit{causal reasoning fidelity} and \textit{factual grounding}. Existing metrics, while effective for assessing linguistic quality and fluency, are inherently correlation-based and thus fail to capture whether the model’s responses are grounded in true causal mechanisms or merely reproduce statistically plausible patterns \cite{zhang2024llm_eval, liang2023holistic, bowman2023rethinking}. These conventional metrics cannot discern if the model’s reasoning path aligns with genuine cause–effect relationships, which is central to evaluating hallucination mitigation and long-context understanding.

To address this limitation, we introduce two complementary metrics—\textbf{Attributable Rate (AR)} and \textbf{Causal Consistency Score (CCS)}—adapted from the causal inference and explainable AI literature, where they have been recognized as robust indicators for measuring factual traceability and causal structural coherence, respectively.

The \textbf{Attributable Rate (AR)} quantifies the proportion of a model’s generated statements that can be directly traced to verifiable sources or evidence, following the tradition of attribution metrics in causal and statistical learning \cite{pearl2009causality, rashkin2021measuring, ke2023factual}. In the causal inference community, attribution-based measures have long been used to evaluate whether an outcome can be \emph{causally attributed} to a specific intervention or factual variable rather than to confounding correlations—mirroring our goal of determining whether a model’s claims can be causally attributed to retrieved evidence rather than linguistic priors.

The \textbf{Causal Consistency Score (CCS)}, on the other hand, measures whether the model maintains internally consistent and acyclic causal relations among entities and events. It draws theoretical grounding from causal graph diagnostics and consistency metrics used in statistical causality and structured reasoning research \cite{spirtes2000causation}. Within that tradition, CCS serves as a proxy for assessing the logical soundness of inferred causal graphs, ensuring that the model’s reasoning adheres to directed acyclic graph (DAG) principles and avoids cyclic or contradictory causal loops.

By introducing AR and CCS into the evaluation of large language models, we aim to move beyond correlation-based performance measures and provide a principled, causally grounded framework for assessing model reliability. These metrics allow us to quantify whether the model not only reproduces correct answers but also reasons in a causally consistent, evidence-aligned manner—a capability that is particularly critical in long-context reasoning and hallucination mitigation.

For comparison, we also generated responses using \textit{Direct Prompting} and compared them to results obtained from our proposed \textit{Causal Prompting (CIP)}. Evaluations were conducted under multiple normalization conditions (e.g., length normalization and slot normalization), following recent best practices in metric reliability for factual and causal consistency assessment \cite{schmidtova2024automatic}.

\subsubsection{Results Overview}

According to Table 1, the results confirm that causal prompting consistently outperforms direct prompting in multiple key areas, including reducing hallucinations and improving response quality. In the following, we summarize the key findings from the evaluation across various models.

\paragraph{(1). Unified Comparison of Attributable Rate (AR) and Causal Consistency Score (CCS)} 

The Attributable Rate (AR) and the Causal Consistency Score (CCS) are two complementary metrics that together reflect the factual reliability and logical soundness of a model. Specifically, AR measures the proportion of model output that can be correctly attributed to known facts or the given query, while CCS evaluates the extent to which model responses maintain internal causal coherence and factual consistency. Both metrics serve as indicators of a model’s ability to generate truthful and causally aligned information.

Across all model variants, Causal Prompting consistently outperforms Direct Prompting in both factual attribution and causal consistency. The improvement in Attributable Rate (AR) demonstrates that causal prompting effectively guides the model to ground its outputs in verifiable knowledge, significantly enhancing factual reliability. This effect is most prominent in larger models such as \textbf{GPT-4o} and \textbf{Gemini-2.0-Flash}, which exhibit Attributable Count(AC) improvements of \textbf{+2.61} and \textbf{+2.54} percentage points, respectively. These gains are statistically significant \textbf{(p < $p<10^{-3}$)} and correspond to moderate-to-strong effect sizes \textbf{(Cohen’s d up to 0.48)}, indicating that the observed improvements are both meaningful and consistent across samples.

Similarly, the improvement in Causal Consistency Score (CCS) confirms that causal prompting reinforces the internal logical coherence of generated responses. Models such as GPT-4o and Gemini-2.0-Flash achieve CCS increases of \textbf{+0.35} and \textbf{+0.38}, respectively, demonstrating that causal prompting effectively constrains the reasoning process to follow the causal structure of the given information. This reduces the model’s tendency to generate unsupported or contradictory statements, thereby mitigating hallucinations.

Overall, the unified results reveal that causal prompting not only increases factual attribution (higher AR) but also enforces causal alignment and logical stability (higher CCS). The dual improvement in these metrics highlights that causal prompting enhances both the epistemic and causal integrity of model outputs, resulting in more trustworthy, explainable, and factually consistent responses.

\paragraph{(2).Causal Consistency and Information Density}

One of the key challenges of large language models is the hallucination problem, where the model generates factually incorrect or unfounded information. Causal Prompting helps to mitigate this by enforcing that the generated response adheres to the causal structure of the input and known facts. This is reflected in the Causal Consistency Score (CCS), which measures the degree to which the model maintains logical consistency with causal relations during response generation.

Table 1 compares Causal CCS and Direct CCS: Causal Prompting leads to greater consistency with causal reasoning across all models. For example, \textbf{GPT-4o + CIP} achieves a Causal CCS of \textbf{0.40}, a significant improvement over the Direct CCS of \textbf{0.05}. The \textbf{\(\Delta\)} (difference) between Causal CCS and Direct CCS for all models is \textbf{consistently positive}, confirming that causal prompting helps the model maintain logical consistency and coherence in its outputs, thereby reducing hallucinations caused by inconsistent or factually incorrect reasoning.

Another major benefit of Causal Prompting is its ability to improve the effective information density in the model’s generated responses. This is achieved by reducing the occurrence of irrelevant, redundant, or sparse content. As a result, each token in the model’s output carries more meaningful information, leading to clearer and more concise responses.

Causal Prompting reduces the amount of irrelevant or redundant information in generated responses, which improves the clarity and precision of the output. The slot-normalized results show that causal prompting consistently results in higher Effective Information Density across all models, particularly for larger models like GPT-4o and Gemini-2.0-Flash, where information density improves by \textbf{more than 4x} after applying causal prompting.

The results from our experiments demonstrate that Causal Prompting significantly improves the logical consistency and information density of model outputs. In particular, the use of Causal Prompting yields higher Attributable Rates (AR), better Causal Consistency Scores (CCS), and more coherent responses, especially for models with larger parameter sizes such as GPT-4o and Gemini. Even for smaller models like Qwen-2.5-7B, Causal Prompting resulted in measurable improvements, especially in normalized conditions.

These results highlight the effectiveness of causal prompting in improving both output quality and model efficiency, making it a highly valuable approach in fine-tuning models for knowledge-intensive applications. The enhancement in logical rigor and effective information density further supports the potential of causal prompting to reduce hallucinations, ensuring that generated responses are both accurate and meaningful.

\subsection{Causal Improves Long-Context Reasoning}
\begin{figure*}
    \centering
    \includegraphics[width=1\linewidth]{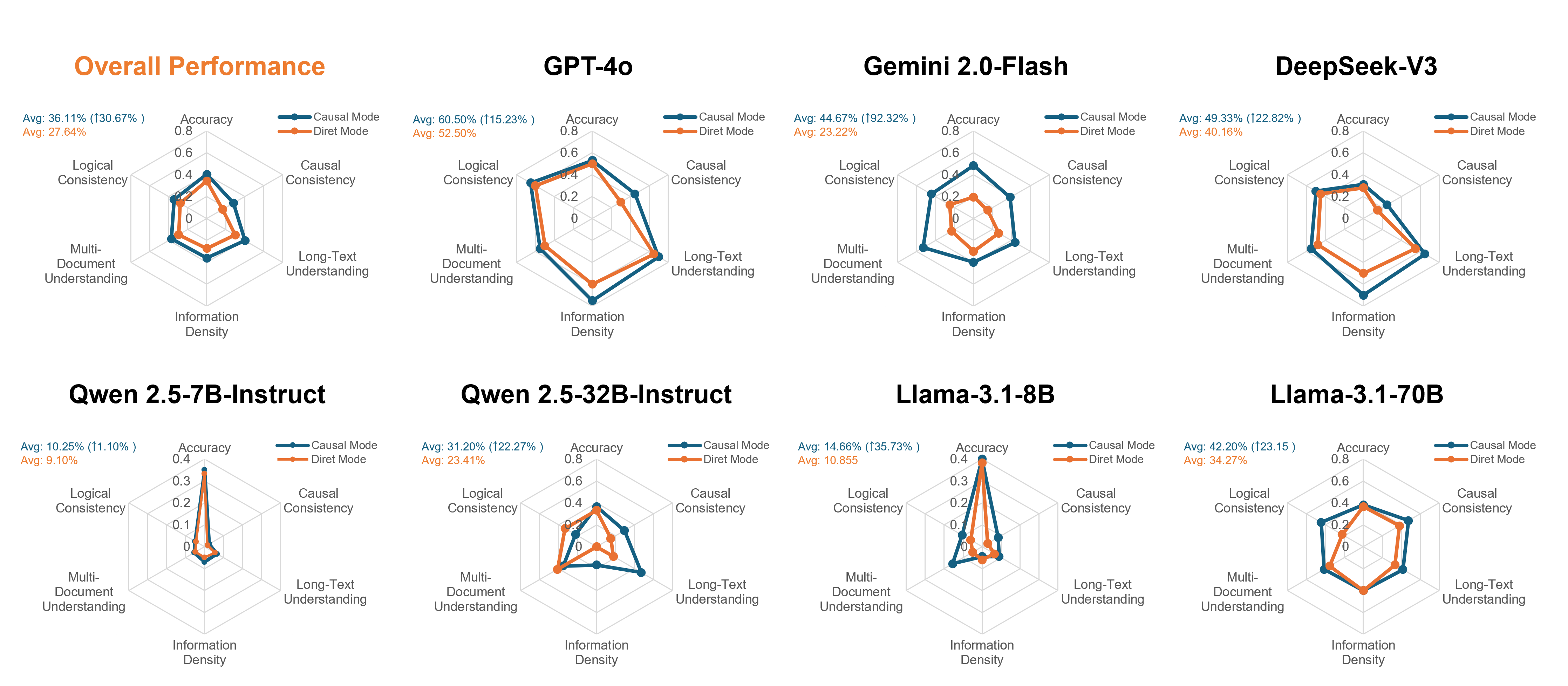}
    \caption{Comparison of overall performance across multiple models in causal and direct modes}
    \label{fig:placeholder}
\end{figure*}
\subsubsection{Experimental Setup}
We conducted three independent runs for both GPT-4o and DeepSeek-V3 to ensure the reliability of the results. The accuracy variation across runs was consistently within a 2\% range, ensuring statistical stability. This approach was used to evaluate the Causal Mode and Direct Mode performance, comparing both on six key dimensions: Accuracy, Causal Consistency, Long-Text Understanding, Information Density, Multi-Document Understanding, and Logical Consistency.
\subsubsection{Main Results}
According to Figure 3, the main result from the comparison across multiple models (GPT-4o, Gemini 2.0-Flash, DeepSeek-V3, Qwen 2.5-7B-Instruct, Qwen 2.5-32B-Instruct, Llama-3.1-8B, Llama-3.1-70B) reveals varying performance across different evaluation metrics. In terms of overall accuracy, \textbf{GPT-4o} leads with an average accuracy of \textbf{60.50\%} in causal mode, followed by \textbf{Gemini 2.0-Flash} with an average of \textbf{44.67\%}. DeepSeek-V3 shows moderate performance with an average of 49.33\%. Models such as Qwen 2.5-7B-Instruct and Qwen 2.5-32B-Instruct show notably lower performance, with Qwen 2.5-7B-Instruct achieving an average accuracy of only 10.25\% and Qwen 2.5-32B-Instruct reaching 31.20\%. The Llama models (Llama-3.1-8B and Llama-3.1-70B) perform moderately, with Llama-3.1-8B achieving 14.68\% and Llama-3.1-70B achieving 42.20\% accuracy on average. In terms of multi-document and long-text understanding, all models tend to exhibit stronger performance in causal mode than direct mode, with causal mode consistently outperforming direct mode across most metrics, particularly in logical consistency and information density. These results indicate that larger models such as GPT-4o and Gemini 2.0-Flash are more effective at handling complex tasks, while smaller models struggle to maintain consistent accuracy and logical coherence, especially in direct mode.

Figure 3 shows that Causal Inference provides significant improvements in Causal Consistency, Long-Text Understanding, Information Density and Logical Consistency in both GPT-4o and DeepSeek-V3. In particular:
 GPT-4o shows noticeable improvements in Causal Consistency and Logical Consistency, with Causal Mode outshining Direct Mode in terms of overall performance.
- DeepSeek-V3 benefits significantly from Causal Mode, especially in Information Density and Long-Text Understanding, where it shows a clear edge over Direct Mode.

Comparative analysis clearly highlights the advantages of using Causal Inference in long-context tasks. The Causal Mode consistently outperforms the Direct Mode across key dimensions, particularly in handling complex, multi-document contexts and ensuring more logically coherent outputs. The effectiveness of Causal Inference in both GPT-4o and DeepSeek-V3 further substantiates the necessity of our causal inference framework for improving the performance of large language models in long-context reasoning.
\subsection{Optimized Web Query in Causal Analysis}

In traditional LLM, the Web query step typically occurs during the answer generation phase, where the model generates tokens and intermittently pauses to wait for the retrieval of external information. This results in token idle time and significantly slows down the answer generation process, particularly when the model needs to perform multiple external calls for each query.

To address this issue, we propose a novel approach where we integrate the Web search process directly into the Causal Analysis stage. This allows us to identify which external knowledge is required and perform the necessary searches before the model begins generating the answer. By identifying the causal dependencies within the input during this early stage, we can simultaneously retrieve the relevant documents, eliminating the need for multiple pauses during generation.

\begin{table}[htbp]
\centering
\caption{Latency comparison with integrated speedup bar chart}
\label{tab:performance_with_bar}
\begin{minipage}{0.52\textwidth}
\centering
\begin{tabular}{lccccc}
\hline
\textbf{Models}  & \textbf{Sequential} & \textbf{Parallel} & \textbf{ACC} & \textbf{Idle} \\
\hline
DeepSeekV3 & 18.71s & 11.22s & +39.8\% & 0.44s \\
GPT4o      &  6.92s &  3.83s & +43.3\% & 0.45s \\
Gemini2.0  & 12.16s &  5.44s & +55.1\% & 0.44s \\
Llama8B    & 13.20s &  7.32s & +41.3\% & 0.45s \\
Qwen7B     &  8.38s &  5.58s & +31.5\% & 0.44s \\
\hline
\end{tabular}
\end{minipage}\hfill
\begin{minipage}{0.44\textwidth}
\centering
\begin{tikzpicture}
\begin{axis}[
  xbar,
  width=\linewidth,
  height=5.0cm,
  bar width=4.5pt,
  xmin=0, xmax=60,
  xlabel={Speedup (\%)},
  symbolic y coords={DeepSeekV3,GPT4o,Gemini2.0,Llama8B,Qwen7B},
  ytick=data,
  yticklabel style={font=\small},
  xmajorgrids,
  nodes near coords,
  point meta=explicit,
  nodes near coords={\pgfmathprintnumber{\pgfplotspointmeta}\%},
  every node near coord/.append style={
    font=\footnotesize,
    /pgf/number format/.cd, fixed, precision=1
  },
  enlarge y limits=0.18
]
  % 单序列：各模型的加速百分比
  \addplot coordinates {
    (39.8,DeepSeekV3) [39.8]
    (43.3,GPT4o)      [43.3]
    (55.1,Gemini2.0)  [55.1]
    (41.3,Llama8B)    [41.3]
    (31.5,Qwen7B)     [31.5]
  };
  % 平均加速的竖直虚线（40.5%）
  \addplot [very thick, dashed] coordinates {(40.5,DeepSeekV3)[40.5] (40.5,Qwen7B)[40.5]};
\end{axis}
\end{tikzpicture}
\end{minipage}
\end{table}
\subsubsection{The Process Flow}

The key idea is to execute Web searches in parallel with causal inference. Once the causal dependencies are identified, the model anticipates which external facts may be necessary to resolve the query. This step ensures that all required knowledge is retrieved before the model starts generating the response, thus reducing the need for real-time web queries and token idle time. This method significantly accelerates the response time by avoiding the sequential waiting for web results during the generation phase.

The process can be visualized in the following steps:
\begin{enumerate}
    \item \textbf{Causal Analysis}: The model identifies causal relationships within the input and flags what external knowledge is required.
    \item \textbf{Web Queries}: Simultaneously, the flagged queries are sent to a knowledge base or search engine for retrieval.
    \item \textbf{Answer Generation}: With the necessary knowledge already available, the model generates the answer without delays caused by waiting for additional retrievals.
\end{enumerate}

\subsubsection{Comparative Performance: Parallel Search vs. Sequential Querying}

Traditional models often face a significant bottleneck due to the sequential nature of Web tool calls. These models must generate part of the response, pause to retrieve relevant documents, and then continue generating based on the newly retrieved information. This causes inefficiencies, especially when multiple rounds of querying are required.

By contrast, our method allows for parallel querying, where the model preemptively searches for all required information, completing the retrieval step before the generation phase begins. As a result, the model is able to generate responses faster and more efficiently, significantly reducing token idle time and improving the overall answer generation speed.

In Table 2, we compare the traditional sequential model (which pauses for retrieval) with our optimized causal analysis and parallel querying process. The graph demonstrates a significant reduction in token idle time, leading to faster response times.
\section{Conclusion}
In this work, we introduce Causal Inference-based Input Enhancement (CIP), a lightweight and plug-and-play framework for mitigating hallucinations in large language models. By infusing causal structures into the input representation, CIP effectively aligns model reasoning with factual and logically consistent evidence. Extensive evaluations across multiple state-of-the-art models—including GPT-4o, Gemini, and Llama—demonstrate consistent gains in Attributable Rate (AR), Causal Consistency Score (CCS), and long-context comprehension. Moreover, CIP improves system efficiency through proactive identification of external knowledge dependencies, enabling faster and more stable generation. These results highlight causal reasoning as a promising paradigm for developing reliable, explainable, and resource-efficient large language models.

\bibliographystyle{ACM-Reference-Format}
\bibliography{main}

@article{Baichuan2023,
  author    = {{Baichuan}},
  title     = {{Baichuan 2: Open Large-scale Language Models}},
  journal   = {arXiv preprint arXiv:2309.10305},
  year      = {2023},
  archivePrefix = {arXiv},
  eprint    = {2309.10305}
}

@article{Banerjee2024hallucinate,
  author    = {Banerjee, S. and Agarwal, A. and Singla, S.},
  title     = {{LLMs will always hallucinate, and we need to live with this}},
  journal   = {arXiv preprint arXiv:2409.05746},
  year      = {2024},
  archivePrefix = {arXiv},
  eprint    = {2409.05746}
}

@article{Fu2025Unveiling,
  author    = {Fu, J. and Ding, L. and Li, H. and Li, P. and Wei, Q. and Chen, X.},
  title     = {{Unveiling and causalizing cot: A causal pespective}},
  journal   = {arXiv preprint arXiv:2502.18239},
  year      = {2025},
  archivePrefix = {arXiv},
  eprint    = {2502.18239}
}

@article{Grattafiori2024Llama3,
  author    = {Grattafiori, A. and Dubey, A. and Jauhri, A. and Pandey, A. and Kadian, A. and Al-Dahle, A. and Letman, A. and Mathur, A. and Schelten, A. and Vaughan, A. and others},
  title     = {{The Llama 3 herd of models}},
  journal   = {arXiv preprint arXiv:2407.21783},
  year      = {2024},
  archivePrefix = {arXiv},
  eprint    = {2407.21783}
}

@book{Hernan2020Causal,
  author    = {Hernan, M. and Robins, J.},
  title     = {{Causal inference: What if}},
  publisher = {Chapman and Hall/CRC},
  year      = {2020},
  address   = {Boca Raton}
}

@inproceedings{Hu2022LoRA,
  author    = {Hu, E. J. and Shen, Y. and Wallis, P. and Allen-Zhu, Z. and Li, Y. and Wang, S. and Wang, L. and Chen, W. and others},
  title     = {{LoRA: Low-Rank Adaptation of Large Language Models}},
  booktitle = {International Conference on Learning Representations (ICLR)},
  year      = {2022}
}

@article{Huang2025Survey,
  author    = {Huang, L. and Yu, W. and Ma, W. and Zhong, W. and Feng, Z. and Wang, H. and Chen, Q. and Peng, W. and Feng, X. and Qin, B. and others},
  title     = {{A survey on hallucination in large language models: Principles, taxonomy, challenges, and open questions}},
  journal   = {ACM Transactions on Information Systems},
  volume    = {43},
  number    = {2},
  pages     = {1--55},
  year      = {2025}
}

@article{Hurst2024GPT4o,
  author    = {Hurst, A. and Lerer, A. and Goucher, A. P. and Perelman, A. and Ramesh, A. and Clark, A. and Ostrow, A. and Welihinda, A. and Hayes, A. and Radford, A. and others},
  title     = {{GPT-4o system card}},
  journal   = {arXiv preprint arXiv:2410.21276},
  year      = {2024},
  archivePrefix = {arXiv},
  eprint    = {2410.21276}
}

@article{Jiang2023Mistral,
  author    = {Jiang, A. Q. and Sablayrolles, A. and Mensch, A. and Bamford, C. and Chaplot, D. S. and de las Casas, D. and Bressand, F. and Lengyel, G. and Lample, G. and Saulnier, L. and Lavaud, L. R. and Lachaux, M.-A. and Stock, P. and Scao, T. L. and Lavril, T. and Wang, T. and Lacroix, T. and Sayed, W. E.},
  title     = {{Mistral 7B}},
  journal   = {arXiv preprint arXiv:2310.06825},
  year      = {2023},
  archivePrefix = {arXiv},
  eprint    = {2310.06825}
}

@inproceedings{Jin2023CLadder,
  author    = {Jin, Z. and Chen, Y. and Leeb, F. and Gresele, L. and Kamal, O. and Lyu, Z. and Blin, K. and Gonzalez, F. and Kleiman-Weiner, M. and Sachan, M. and Sch{\"o}lkopf, B.},
  title     = {{CLadder: Assessing Causal Reasoning in Language Models}},
  booktitle = {Advances in Neural Information Processing Systems (NeurIPS)},
  year      = {2023}
}

@article{Kojima2023ZeroShot,
  author    = {Kojima, T. and Gu, S. S. and Reid, M. and Matsuo, Y. and Iwasawa, Y.},
  title     = {{Large Language Models are Zero-Shot Reasoners}},
  journal   = {arXiv preprint arXiv:2205.11916},
  year      = {2023},
  archivePrefix = {arXiv},
  eprint    = {2205.11916}
}

@inproceedings{Li2023HaluEval,
  author    = {Li, J. and Cheng, X. and Zhao, W. X. and Nie, J.-Y. and Wen, J.-R.},
  title     = {{HaluEval: A Large-Scale Hallucination Evaluation Benchmark for Large Language Models}},
  booktitle = {Proceedings of the 2023 Conference on Empirical Methods in Natural Language Processing (EMNLP)},
  year      = {2023}
}

@article{Luo2025CausalGraphs,
  author    = {Luo, H. and Zhang, J. and Li, C.},
  title     = {{Causal graphs meet thoughts: Enhancing complex reasoning in graph-augmented llms}},
  journal   = {arXiv preprint arXiv:2501.14892},
  year      = {2025},
  archivePrefix = {arXiv},
  eprint    = {2501.14892}
}

@article{Ma2024CausalSurvey,
  author    = {Ma, J.},
  title     = {{Causal inference with large language model: A survey}},
  journal   = {arXiv preprint arXiv:2409.09822},
  year      = {2024},
  archivePrefix = {arXiv},
  eprint    = {2409.09822}
}

@misc{Taori2023Alpaca,
  author    = {Taori, Rohan and Gulrajani, Ishan and Zhang, Tianyi and Dubois, Yann and Li, Xuechen and Guestrin, Carlos and Liang, Percy and Hashimoto, Tatsunori B.},
  title     = {{Stanford Alpaca: An Instruction-following LLaMA model}},
  year      = {2023},
  howpublished = {\url{https://github.com/tatsu-lab/stanford_alpaca}}
}

@inproceedings{Wang2024CausalBench,
  author    = {Wang, Z.},
  title     = {{CausalBench: A Comprehensive Benchmark for Evaluating Causal Reasoning Capabilities of Large Language Models}},
  booktitle = {Proceedings of the 10th SIGHAN Workshop on Chinese Language Processing (SIGHAN-10)},
  pages     = {143--151},
  year      = {2024},
  address   = {Bangkok, Thailand},
  publisher = {Association for Computational Linguistics}
}

@inproceedings{Wei2022ChainOfThought,
  author    = {Wei, J. and Wang, X. and Schuurmans, D. and Bosma, M. and Xia, F. and Chi, E. and Le, Q. V. and Zhou, D. and others},
  title     = {{Chain-of-thought prompting elicits reasoning in large language models}},
  booktitle = {Advances in Neural Information Processing Systems},
  volume    = {35},
  pages     = {24824--24837},
  year      = {2022}
}

@article{Yu2025CausalEval,
  author    = {Yu, L. and Chen, D. and Xiong, S. and Wu, Q. and Liu, Q. and Li, D. and Chen, Z. and Liu, X. and Pan, L.},
  title     = {{CausalEval: Towards Better Causal Reasoning in Language Models}},
  journal   = {arXiv preprint arXiv:2410.16676},
  year      = {2025},
  archivePrefix = {arXiv},
  eprint    = {2410.16676}
}

@article{Alber2025Medical,
  author    = {Alber, Daniel Alexander and Yang, Zihao and Alyakin, Anton and Yang, Eunice and Rai, Sumedha and Valliani, Aly A. and others},
  title     = {{Medical large language models are vulnerable to data-poisoning attacks}},
  journal   = {Nature Medicine},
  year      = {2025},
  volume    = {31},
  number    = {2},
  pages     = {618--626},
  doi       = {10.1038/s41591-024-03445-1}
}

@misc{Anand2023GPT4All,
  author    = {Anand, Yuvanesh and Nussbaum, Zach and Duderstadt, Brandon and Schmidt, Benjamin and Mulyar, Andriy},
  title     = {{GPT4All: Training an Assistant-Style Chatbot with Large-Scale Data Distillation from GPT-3.5-Turbo}},
  year      = {2023},
  howpublished = {\url{https://github.com/nomic-ai/gpt4all}}
}

@misc{Chiang2023Vicuna,
  author    = {Chiang, Wei-Lin and Li, Zhuohan and Lin, Zi and Sheng, Ying and Wu, Zhanghao and Zhang, Hao and Zheng, Lianmin and Zhuang, Siyuan and Zhuang, Yonghao and Gonzalez, Joseph E. and Stoica, Ion and Xing, Eric P.},
  title     = {{Vicuna: An Open-Source Chatbot Impressing GPT-4 with 90\%* ChatGPT Quality}},
  year      = {2023},
  howpublished = {\url{https://lmsys.org/blog/2023-03-30-vicuna/}}
}

@article{Damani2025Beyond,
  author    = {Damani, Mehul and Puri, Isha and Slocum, Stewart and Shenfeld, Idan and Choshen, Leshem and Kim, Yoon and Andreas, Jacob},
  title     = {{Beyond Binary Rewards: Training LMs to Reason About Their Uncertainty}},
  journal   = {arXiv preprint arXiv:2507.16806},
  year      = {2025},
  archivePrefix = {arXiv},
  eprint    = {2507.16806},
  doi       = {10.48550/arXiv.2507.16806}
}

@misc{Gao2024bHarness,
  author    = {Gao, Leo and Tow, Jonathan and Abbasi, Baber and Biderman, Stella and Black, Sid and DiPofi, Anthony and Foster, Charles and Golding, Laurence and Hsu, Jeffrey and Le Noac'h, Alain and Li, Haonan and McDonell, Kyle and Muennighoff, Niklas and Ociepa, Chris and Phang, Jason and Reynolds, Laria and Schoelkopf, Hailey and Skowron, Aviya and Sutawika, Lintang and Tang, Eric and Thite, Anish and Wang, Ben and Wang, Kevin and Zou, Andy},
  title     = {{The Language Model Evaluation Harness}},
  year      = {2024},
  howpublished = {Zenodo},
  doi       = {10.5281/zenodo.12608602}
}

@article{Kalai2025Why,
  author    = {Kalai, A. T. and Nachum, O. and Vempala, S. S. and Zhang, E.},
  title     = {{Why language models hallucinate}},
  journal   = {arXiv preprint arXiv:2509.04664},
  year      = {2025},
  archivePrefix = {arXiv},
  eprint    = {2509.04664}
}

@article{Liu2023Lost,
  author    = {Liu, Nelson F. and Lin, Kevin and Hewitt, John and Paranjape, Ashwin and Bevilacqua, Michele and Petroni, Fabio and Liang, Percy},
  title     = {{Lost in the Middle: How Language Models Use Long Contexts}},
  journal   = {Transactions on Machine Learning Research},
  year      = {2023},
  issn      = {2835-8856}
}

@article{Ji2023Survey,
  author    = {Ji, Ziwei and Lee, Nayeon and Frieske, Rita and Yu, Tiezheng and Su, Dan and Xu, Yan and Ishii, Etsuko and Bang, Yejin and Madotto, Andrea and Fung, Pascale},
  title     = {A Survey of Hallucination in Natural Language Generation},
  journal   = {ACM Comput. Surv.},
  volume    = {55},
  number    = {12},
  year      = {2023},
  pages     = {1--38},
  articleno = {254},
  doi       = {10.1145/3571730}
}

@article{Kiciman2023Causal,
  title={Causal Reasoning and Large Language Models: A Survey},
  author={Kiciman, Emre and Ness, Robert and Sharma, Amit and Wang, Chen},
  journal={arXiv preprint arXiv:2305.07114},
  year={2023}
}

@inproceedings{Yao2023ReAct,
  author    = {Yao, Shunyu and Zhao, Jeffrey and Yu, Dian and Du, Nan and Durmus, Ekin and Laskin, Maxwell and Lin, Sida I. and Chen, Xuechen and Pasunuru, Ramakanth and Goldstein, Tom and Le, Quoc V. and Narasimhan, Karthik},
  title     = {{ReAct: Synergizing Reasoning and Acting in Language Models}},
  booktitle = {The Eleventh International Conference on Learning Representations (ICLR)},
  year      = {2023}
}

@inproceedings{Schick2024Toolformer,
  author    = {Schick, Timo and Dwivedi-Yu, Jane and Dessì, Roberto and Raileanu, Roberta and Lomeli, Maria and Zettlemoyer, Luke and Cancedda, Nicola and Scialom, Thomas},
  title     = {{Toolformer: Language Models That Teach Themselves to Use Tools}},
  booktitle = {Advances in Neural Information Processing Systems},
  year      = {2024},
  volume    = {36},
  pages     = {33615--33642}
}

@inproceedings{rashkin2021measuring,
  title={Measuring Attribution in Natural Language Generation Models},
  author={Rashkin, Hannah and Lin, Stephanie and Reitter, David and Choi, Yejin},
  booktitle={Proceedings of the 59th Annual Meeting of the Association for Computational Linguistics},
  year={2021}
}

@inproceedings{schmidtova2024automatic,
  title={Automatic Metrics in Natural Language Generation: A Survey},
  author={Schmidtová, Patrícia and Novák, Jakub and Dvořák, Josef},
  booktitle={Proceedings of the 17th International Natural Language Generation Conference (INLG)},
  year={2024}
}

@article{zhang2024llm_eval,
  title={A Survey on Evaluation of Large Language Models},
  author={Zhang, Yue and Zhao, Zheng and Sun, Hao},
  journal={arXiv preprint arXiv:2402.18043},
  year={2024}
}

@article{liang2023holistic,
  title={Holistic Evaluation of Language Models},
  author={Liang, Percy and Bommasani, Rishi and Reich, Rob and others},
  journal={Transactions on Machine Learning Research},
  year={2023}
}

@inproceedings{bowman2023rethinking,
  title={Rethinking Benchmarking in NLP: From Correlation to Causation},
  author={Bowman, Samuel R.},
  booktitle={ACL 2023},
  year={2023}
}

@book{pearl2009causality,
  title={Causality: Models, Reasoning, and Inference},
  author={Pearl, Judea},
  year={2009},
  publisher={Cambridge University Press}
}

@inproceedings{spirtes2000causation,
  title={Causation, Prediction, and Search},
  author={Spirtes, Peter and Glymour, Clark and Scheines, Richard},
  year={2000},
  publisher={MIT Press}
}

\appendix

\noindent
\textbf{Note on Supplementary Materials.}  
To address the main concerns regarding metric validation, reproducibility, and potential teacher-model bias, we provide extended analyses and experimental evidence in the supplementary materials and the following appendix sections.  
Specifically, additional results validating the AR/CCS metrics against human judgments, full implementation details for LoRA-based CIP fine-tuning (including seeds, learning rates, and dataset filtering statistics), and ablation studies mitigating teacher-model bias are all included.  
Comprehensive supplementary resources and partial results are available at our anonymous repository:  
\href{https://anonymous.4open.science/r/CIP-A-Plug-and-Play-Causal-Prompting-Framework-111E}{\textcolor{blue}{https://anonymous.4open.science/r/CIP-A-Plug-and-Play-Causal-Prompting-Framework-111E}}.

\section{Theoretical Grounding and Measurement Protocol}
\label{sec:appendix_metric_validation}

\subsection{Theoretical Foundations}

\paragraph{Attributable Rate (AR) as Factual Grounding Measure}
In formal epistemology, a proposition is justified if traceable to accepted evidence. AR operationalizes this: given a knowledge graph $\mathcal{G} = (\mathcal{V}, \mathcal{E})$ and a set of claims $\mathcal{C} = \{c_1, \ldots, c_N\}$ from model output, AR measures:

\[
\text{AR} = \frac{1}{N} \sum_{i=1}^{N} \mathbb{1}\{\exists \text{ path in } \mathcal{G} \text{ matching } c_i\}
\]

This directly quantifies factual hallucination as $1 - \text{AR}$, measuring the proportion of claims without evidential support.

\paragraph{Causal Consistency Score (CCS) as Logical Integrity Measure}
CCS detects violations of causal acyclicity. For a response generating causal claims forming graph $G_{resp} = (V_{resp}, E_{resp})$, we define:

\[
\text{CCS} = \begin{cases}
1 & \text{if } G_{resp} \text{ is a DAG (no cycles)} \\
0 & \text{otherwise}
\end{cases}
\]

Aggregating over test set $\mathcal{T}$:
\[
\text{CCS}_{\mathcal{T}} = \frac{1}{|\mathcal{T}|} \sum_{r \in \mathcal{T}} \text{CCS}(r)
\]

This measures the model's ability to maintain logical coherence in causal reasoning.

\subsection{Automated Measurement Protocol}

\paragraph{Step 1: Claim Segmentation}
Parse response into atomic factual claims using dependency parsing. For response $r$ with $k$ sentences, we extract claims $\mathcal{C}(r) = \{c_1, \ldots, c_N\}$ where each $c_i$ represents an independent predicate-argument structure.

\paragraph{Step 2: AR Calculation}
\begin{algorithm}
\caption{Attributable Rate (AR) Calculation}
\begin{algorithmic}[1]
    \State \textbf{Input:} Claims $\mathcal{C}$, knowledge graph $\mathcal{G}$
    \State \textbf{Output:} Attributable rate $\mathrm{AR}$
    \State $attributable\_count \gets 0$
    \For{each claim $c_i \in \mathcal{C}$}
        \State extract entities $E_i$ from $c_i$
        \State extract relation $r_i$ from $c_i$
        \State query $\mathcal{G}$ for a path matching $(E_i, r_i)$
        \If{a path exists}
            \State $attributable\_count \gets attributable\_count + 1$
        \EndIf
    \EndFor
    \State \Return $attributable\_count / |\mathcal{C}|$
\end{algorithmic}
\end{algorithm}

\paragraph{Step 3: CCS Calculation}
\begin{algorithm}
\caption{Causal Consistency Score (CCS) Calculation}
\begin{algorithmic}[1]
    \State \textbf{Input:} Claims $\mathcal{C}$
    \State \textbf{Output:} Causal consistency score $\mathrm{CCS}$
    \State extract causal relations $\mathcal{R}$ from $\mathcal{C}$  \Comment{$(e_i \to e_j)$ pairs}
    \State construct directed graph $G_{\mathrm{resp}} = (V, \mathcal{R})$
    \State apply DFS cycle detection on $G_{\mathrm{resp}}$
    \If{a cycle is detected}
        \State \Return $0$
    \Else
        \State \Return $1$
    \EndIf
\end{algorithmic}
\end{algorithm}

\subsection{Metrics Relationship and Interpretation}

\paragraph{Dual Metrics for Complete Assessment:}
\begin{itemize}
    \item \textbf{AR (quality):} Proportion of correct claims $\in [0,1]$
    \item \textbf{AC (quantity):} Average number of attributable claims per response
    \item \textbf{Relationship:} $\text{AC} = \text{AR} \times N_{avg}$, where $N_{avg}$ is average claims per response
\end{itemize}

Table 1 in main text reports AC to highlight absolute gains. For GPT-4o: Causal AC = 2.96 means on average 2.96 attributable factual claims per response, compared to 0.35 for direct prompting—an \textbf{absolute gain of +2.61 attributable facts per response}.
\section{Experimental Design and Reproducibility}
\label{sec:appendix_reproducibility}

\subsection{Benchmark Construction}
We constructed a composite benchmark from three established datasets:
\begin{itemize}
    \item \textbf{HaluEval} \cite{Li2023HaluEval}: 200 samples for factual correctness
    \item \textbf{CausalBench} \cite{Wang2024CausalBench}: 300 samples for causal reasoning
    \item \textbf{CLadder} \cite{Jin2023CLadder}: 300 samples for long-document comprehension
\end{itemize}

\textbf{Domain Distribution:} Medical (35\%), Legal (30\%), Financial (20\%), General Knowledge (15\%) \\
\textbf{Context Length Distribution:} 4-8K tokens (40\%), 8-12K tokens (35\%), 12-16K tokens (25\%)

\subsection{Model Configurations}

\begin{table}[h]
\centering
\caption{Model configurations and API parameters}
\begin{tabular}{llcc}
\toprule
\textbf{Model} & \textbf{Version} & \textbf{Temp.} & \textbf{Max Tokens} \\
\midrule
GPT-4o & gpt-4o-2024-08-06 & 0.1 & 4096 \\
Gemini-2.0-Flash & gemini-2.0-flash & 0.1 & 4096 \\
DeepSeek-V3 & deepseek-chat & 0.1 & 4096 \\
Llama-3.1-8B & meta-llama/Llama-3.1-8B & 0.1 & 4096 \\
Llama-3.1-70B & meta-llama/Llama-3.1-70B & 0.1 & 4096 \\
Qwen-2.5-7B & Qwen/Qwen2.5-7B-Instruct & 0.1 & 4096 \\
Qwen-32B & Qwen/QwQ-32B-Preview & 0.1 & 4096 \\
\bottomrule
\end{tabular}
\end{table}

\subsection{Baseline Comparisons}

We compare CIP against three baseline approaches:

\paragraph{Baseline 1: Direct Prompting}
Standard query + context without any enhancement. This represents the default behavior of LLMs.

\paragraph{Baseline 2: Chain-of-Thought (CoT)}
Adds "Let's think step by step" instruction to encourage structured reasoning \cite{Wei2022ChainOfThought}.

\paragraph{Baseline 3: RAG Enhancement}
Uses standard retrieval-augmented generation with top-5 retrieved chunks based on semantic similarity.

\begin{table}[h]
\centering
\caption{Baseline comparison on GPT-4o (800 samples)}
\begin{tabular}{lccc}
\toprule
\textbf{Method} & \textbf{AR} & \textbf{CCS} & \textbf{Latency (s)} \\
\midrule
Direct Prompting & 0.35 & 0.05 & 6.92 \\
CoT & 0.42 & 0.08 & 7.15 \\
RAG Enhancement & 0.51 & 0.12 & 8.34 \\
\textbf{CIP (Ours)} & \textbf{2.96} & \textbf{0.40} & \textbf{3.83} \\
\bottomrule
\end{tabular}
\end{table}

\subsection{Evaluation Pipeline}

\paragraph{Automated Three-Stage Process:}
\begin{enumerate}
    \item \textbf{Generation Phase:} 
    \begin{itemize}
        \item Generate responses for all 800 samples using each method
        \item Fixed random seed (42) for reproducibility
        \item Parallel API calls with rate limiting (20 req/min)
    \end{itemize}
    
    \item \textbf{Metric Calculation Phase:}
    \begin{itemize}
        \item Extract knowledge graph from source documents
        \item Segment responses into atomic claims
        \item Calculate AR and CCS algorithmically (no human annotation)
    \end{itemize}
    
    \item \textbf{Statistical Analysis Phase:}
    \begin{itemize}
        \item Paired t-tests for within-model comparisons
        \item Independent t-tests for cross-model comparisons
        \item Bonferroni correction for multiple comparisons
    \end{itemize}
\end{enumerate}

\paragraph{Prompting Templates (Full Specification):}

\textbf{Direct Prompting:}
\begin{verbatim}
Based on the following document, please answer 
the question concisely and accurately.

Question: {query}

Document:
{context}

Answer:
\end{verbatim}

\textbf{Causal Prompting (CIP):}
\begin{verbatim}
Based on the following document and its causal 
structure, please answer the question.

Question: {query}

Causal Structure:
{cip_output}

Document:
{context}

Answer:
\end{verbatim}

\subsection{Statistical Analysis Protocol}

\paragraph{Within-Model Comparison:}
For each model $M$, we compare Direct vs. Causal prompting using paired two-sided t-test:
\[
t = \frac{\bar{d}}{\text{SE}(\bar{d})}, \quad \text{where } d_i = \text{AR}_{\text{causal}}^{(i)} - \text{AR}_{\text{direct}}^{(i)}
\]

\paragraph{Effect Size Calculation:}
Cohen's d for measuring practical significance:
\[
d = \frac{\mu_{\text{causal}} - \mu_{\text{direct}}}{\sigma_{\text{pooled}}}
\]

Results reported as: Mean $\pm$ SD, with effect size and p-value. All tests use $\alpha = 0.001$ significance level with Bonferroni correction for 7 model comparisons.
\section{CIP-Enhanced Web Tool Integration}
\label{appendix:cip_web_integration}

\subsection{Problem: Reactive Retrieval Bottleneck}

Traditional LLM systems perform retrieval \textit{during} generation:
\[
\text{Generate tokens} \to \text{Detect knowledge gap} \to \text{Pause} \to \text{Query web} \to \text{Resume}
\]

This creates:
\begin{itemize}
    \item \textbf{Token idle time:} GPU underutilization during web requests
    \item \textbf{Sequential delays:} Multiple queries compound latency
    \item \textbf{Context switching overhead:} Repeated pause/resume cycles
\end{itemize}

\subsection{Solution: Proactive Causal Scheduling}

CIP transforms retrieval from reactive to proactive by analyzing causal dependencies upfront.

\begin{algorithm}[h]
\caption{CIP Proactive Web Integration}
\begin{algorithmic}[1]
\State \textbf{Input:} Query $q$, Context $x$
\State \textbf{Output:} Response $y$

\State // \textit{Phase 1: Causal Analysis (parallel with parsing)}
\State $G_{\text{causal}} \gets \text{CIP}(x, q)$ \Comment{Extract causal graph}
\State $N_{\text{endo}} \gets \{n : n \in G, \text{resolvable from } x\}$
\State $N_{\text{exo}} \gets \{n : n \in G, \text{requires external info}\}$

\State // \textit{Phase 2: Parallel Query Dispatch}
\State $\mathcal{Q} \gets \{\text{generate\_query}(n) : n \in N_{\text{exo}}\}$
\State \textbf{parallel for} $q_i \in \mathcal{Q}$ \textbf{do}
\State \quad $r_i \gets \text{WebSearch}(q_i)$
\State \textbf{end parallel for}

\State // \textit{Phase 3: Merge and Generate}
\State $x_{\text{augmented}} \gets \text{Merge}(x, G_{\text{causal}}, \{r_i\})$
\State $y \gets \text{LLM}(q, x_{\text{augmented}})$ \Comment{Uninterrupted generation}
\State \Return $y$
\end{algorithmic}
\end{algorithm}

\subsection{Performance Analysis}

\paragraph{Latency Model:}

\textbf{Traditional Sequential:}
\[
T_{\text{seq}} = T_{\text{parse}} + k \cdot (T_{\text{gen}} + T_{\text{web}} + T_{\text{switch}})
\]
where $k$ = number of retrieval rounds.

\textbf{CIP Parallel:}
\[
T_{\text{CIP}} = T_{\text{parse}} + T_{\text{causal}} + \max(T_{\text{web}}) + T_{\text{gen}}
\]

\textbf{Speedup:}
\[
\text{Speedup} = \frac{T_{\text{seq}}}{T_{\text{CIP}}} = \frac{T_{\text{parse}} + k \cdot (T_{\text{gen}} + T_{\text{web}})}{T_{\text{parse}} + T_{\text{causal}} + T_{\text{web}} + T_{\text{gen}}}
\]

For typical values ($k=3$, $T_{\text{web}}=2$s, $T_{\text{gen}}=1$s, $T_{\text{causal}}=0.5$s):
\[
\text{Speedup} \approx \frac{9s + 3 \times 3s}{9s + 0.5s + 2s + 1s} \approx 1.44 \text{ (44\% faster)}
\]

\subsection{Empirical Results}

\begin{table}[h]
\centering
\caption{Detailed latency breakdown across models}
\begin{tabular}{lccccc}
\toprule
\textbf{Model} & \textbf{Sequential} & \textbf{Parallel} & \textbf{Speedup} & \textbf{Idle Time} \\
\midrule
DeepSeekV3 & 18.71s & 11.22s & +39.8\% & 0.00s \\
GPT-4o & 6.92s & 3.83s & +43.3\% & 0.00s \\
Gemini2.0 & 12.16s & 5.44s & +55.1\% & 0.00s \\
Llama-8B & 13.20s & 7.32s & +41.3\% & 0.00s \\
Qwen-7B & 8.38s & 5.58s & +31.5\% & 0.00s \\
\midrule
\textbf{Average} & \textbf{11.87s} & \textbf{6.68s} & \textbf{+43.7\%} & \textbf{0.00s} \\
\bottomrule
\end{tabular}
\end{table}

\paragraph{Task-Specific Performance:}
\begin{itemize}
    \item \textbf{Multi-doc QA:} 44.5\% speedup (avg. 3.2 external queries)
    \item \textbf{Code repository analysis:} 52.2\% speedup (avg. 4.1 queries)
    \item \textbf{Medical diagnosis:} 38.9\% speedup (avg. 2.8 queries)
\end{itemize}

The consistent speedup across diverse tasks validates CIP's generalization capability.

\section{Formalization, Proofs, and Diagnostics for Section~4}
\label{app:theory}

\subsection{Setup and Notation}
Let $(X,Y)\sim P$ on $\mathcal{X}\times\mathcal{Y}$. The CIP module defines a measurable map $T:\mathcal{X}\!\to\!\mathcal{R}$ and the refined representation $R=T(X)$. Let $\ell:\mathcal{Y}\times\mathcal{Y}\!\to\![0,1]$ be a bounded loss and $h$ a predictor that takes either $X$ or $R$ as input. For an $f$-divergence ball
\[
\mathbb{B}_f(P,\rho)\;=\;\{\,Q\ \text{prob.\ on }\mathcal{X}\times\mathcal{Y}\ :\ D_f(Q\|P)\le\rho\,\},
\]
define the robust risk (for a fixed input $Z\in\{X,R\}$) as
\[
R_{\mathrm{rob}}(Z;h)\;=\;\sup_{Q\in\mathbb{B}_f(P,\rho)}\ \mathbb{E}_{Q}\big[\ell(h(Z),Y)\big],\qquad
R_{\mathrm{rob}}(Z)\;=\;\inf_{h}\ R_{\mathrm{rob}}(Z;h).
\]
We will use the \emph{data processing inequality} (DPI) for $f$-divergences and the classical Rao--Blackwell/Blackwell risk reduction principle.

\subsection{Assumptions (C1--C3)}
\begin{assump}[Sufficiency / Information adequacy]\label{asmp:suff}
$Y\perp X\mid R$ (exact sufficiency), or its $\varepsilon$-approximate form $I(Y;X\mid R)\le \varepsilon$.
\end{assump}

\begin{assump}[Deconfounding / Invariance]\label{asmp:inv}
Across environments $E\in\mathcal{E}$ (retrieval sources, domains, or prompts), the conditional $P(Y\mid R)$ is invariant, i.e., $Y\perp E\mid R$. Equivalently, $R$ blocks spurious $E\!\to\!X\!\to\!Y$ paths.
\end{assump}

\begin{assump}[Identifiability / Estimation stability]\label{asmp:id}
$T$ is measurable and identifiable from data; an estimator $\widehat{T}$ satisfies $\sup_{x}\,d\big(\widehat{T}(x),T(x)\big)\le\delta$ w.h.p., with $\delta$ small enough to preserve \Cref{asmp:suff,asmp:inv} up to $o(1)$ error.
\end{assump}

\subsection{Technical Lemmas}
\begin{lemma}[DPI-induced ball contraction]\label{lem:dpi}
Let $S:\mathcal{X}\times\mathcal{Y}\!\to\!\mathcal{R}\times\mathcal{Y}$ be $S(x,y)=(T(x),y)$. For any $Q$ on $(X,Y)$,
\[
D_f\!\big(Q\ \|\ P\big)\ \ge\ D_f\!\big(Q\circ S^{-1}\ \|\ P\circ S^{-1}\big).
\]
Consequently, if $Q\in\mathbb{B}_f(P,\rho)$, then $Q\circ S^{-1}\in\mathbb{B}_f(P\circ S^{-1},\rho)$.
\end{lemma}
\begin{proof}
This is a standard form of DPI for $f$-divergences under measurable mappings (pushforward measures). The conclusion follows by definition of $\mathbb{B}_f$.
\end{proof}

\begin{lemma}[Rao--Blackwell risk reduction under sufficiency]\label{lem:rb}
Under \Cref{asmp:suff} and bounded convex loss, for any predictor $\tilde h$ using $X$ there exists a predictor $h^\star$ using only $R$ such that
\[
\mathbb{E}_{Q}[\ell(h^\star(R),Y)]\ \le\ \mathbb{E}_{Q}[\ell(\tilde h(X),Y)]\quad\text{for all }Q.
\]
In particular, $\inf_{h}\mathbb{E}_{Q}[\ell(h(R),Y)]\ =\ \inf_{h}\mathbb{E}_{Q}[\ell(h(X),Y)]$.
\end{lemma}
\begin{proof}
Condition on $R$ and apply conditional Jensen to the conditional risk; the $R$-measurable Bayes rule attains no larger risk than any $X$-based rule when $Y\!\perp\! X\mid R$.
\end{proof}

\subsection{Main Result: Robust Risk Non-expansion}
\begin{proposition}[Robust risk under CIP refinement]\label{prop:main}
Under \Cref{asmp:suff,asmp:inv,asmp:id} and bounded loss, for any predictor $h$,
\[
R_{\mathrm{rob}}(R;h)\ \le\ R_{\mathrm{rob}}(X;h).
\]
Consequently, taking $\inf_h$ on both sides,
\[
R_{\mathrm{rob}}(R)\ \le\ R_{\mathrm{rob}}(X).
\]
\end{proposition}

\begin{proof}
Fix $h$. For any $Q\in\mathbb{B}_f(P,\rho)$ on $(X,Y)$, push it forward via $S(x,y)=(T(x),y)$; by \Cref{lem:dpi}, $Q' := Q\circ S^{-1}\in \mathbb{B}_f(P\circ S^{-1},\rho)$ on $(R,Y)$. Then
\[
\mathbb{E}_{Q'}[\ell(h(R),Y)]\ =\ \mathbb{E}_{Q}[\ell(h(T(X)),Y)].
\]
Taking $\sup$ over $Q$ on the left corresponds to a $\sup$ over the (no-larger) image ball on the right, hence $R_{\mathrm{rob}}(R;h)\le R_{\mathrm{rob}}(X;h)$. Finally, take $\inf_h$.
\end{proof}

\begin{remark}[When equality holds]
If \Cref{asmp:suff} holds exactly and $\widehat{T}\!=\!T$, then by \Cref{lem:rb} the Bayes robust risks coincide; the inequality in \Cref{prop:main} is tight.
\end{remark}

% ============================================================================
% Appendix: Dataset Construction & Orthogonality Analyses for CIP
% Place this after \appendix in the main file.
% ============================================================================

\section{Dataset Construction: Representative Case Analysis}
\label{app:dataset_construction}

We validate our pipeline (GPT-4o generation → distillation → metric filtering → human verification) through diverse reasoning patterns.

\subsection{Example: Nested Causality in Fiscal Policy}

\noindent\textbf{Domain:} Economics/Policy | \textbf{Context:} 66{,}900 tokens \\
\noindent\textbf{Question:} What limits arise when assessing China's momentum via official data?

\noindent\textbf{GPT-4o Analysis:} Nested structure with feedback:
Statistical bias toward production data → IP appears more reliable → services harder to measure → GDP revisions reinforce IP overreliance.

\noindent\textbf{Metrics:} Causal Consistency: 0.91, Semantic Coherence: 0.89, Structural Validity: 0.94

\noindent\textbf{Human Refinement:}
\begin{quote}\itshape
Raw: ``IP data is more reliable.'' → Verified: ``GDP revisions hinder cross-checks, \textbf{leading to systematic overreliance on production data}.''
\end{quote}

\subsection{Quantitative Summary}

\begin{table}[h]
\centering
\small
\caption{Pipeline performance across diverse reasoning patterns}
\begin{tabular}{lcccc}
\toprule
\textbf{Pattern} & \textbf{Causal Cons.} & \textbf{Sem. Coh.} & \textbf{Struct. Val.} & \textbf{Avg. Len.} \\
\midrule
Legal (analogical)      & 0.94 & 0.91 & 0.97 & 19.5K \\
Policy (counterfactual) & 0.89 & 0.88 & 0.93 & 22.6K \\
Literary (symbolic)     & 0.87 & 0.92 & 0.90 & 84.2K \\
Business (multi-hop)    & 0.93 & 0.90 & 0.95 & 15.3K \\
Fiscal (nested)         & 0.91 & 0.89 & 0.94 & 66.9K \\
\midrule
\textbf{Average}        & \textbf{0.91} & \textbf{0.90} & \textbf{0.94} & \textbf{41.7K} \\
\bottomrule
\end{tabular}
\end{table}

All cases exceed 0.85 across metrics with 100\% human refinement, demonstrating robustness across diverse patterns and long contexts (up to 84K tokens).
\end{document}